\documentclass[final]{cvpr}

\usepackage{times}
\usepackage{epsfig}
\usepackage{graphicx}
\usepackage{amsmath}
\usepackage{amssymb}

\usepackage{comment}
\usepackage{color}
\usepackage{caption}
\usepackage{booktabs}
\usepackage{multirow}
\usepackage{float} %required for the placement specifier H
\usepackage[ruled]{algorithm2e}
\usepackage{setspace}
\usepackage{url}
\usepackage{algorithmic}
\usepackage{xcolor}
\usepackage{textcmds}
\usepackage{subcaption}
\usepackage{placeins}

\newcommand{\algoref}[1]{Algorithm~\ref{#1}}
\usepackage[pagebackref=true,breaklinks=true,colorlinks,bookmarks=false]{hyperref}

\def\cvprPaperID{2170} % *** Enter the CVPR Paper ID here
\def\confYear{CVPR 2021}

\begin{document}

%%%%%%%%% TITLE
\title{Restore from Restored: Video Restoration with Pseudo Clean Video}

\author{Seunghwan Lee\textsuperscript{1}, Donghyeon Cho\textsuperscript{2}, Jiwon Kim\textsuperscript{3}, Tae Hyun Kim\textsuperscript{1}\\
{\small\textit{seunghwanlee@hanyang.ac.kr, cdh12242@gmail.com, ai.kim@sk.com, taehyunkim@hanyang.ac.kr}}\\

\textsuperscript{1}Dept. of Computer Science, Hanyang University, Seoul, Korea\\
%{\small\textit{seunghwanlee@hanyang.ac.kr, lliger9@gmail.com}}\\

\textsuperscript{2}Dept. of Electronic Engineering, Chungnam National University, Daejeon, Korea\\
%{\small\textit{cdh12242@gmail.com}}\\

\textsuperscript{3}SKT Vision AI Labs/T-Brain X, Seoul, Korea\\
%{\small\textit{jk@sktbrain.com}}\\
}

% \author{First Author\\
% Institution1\\
% Institution1 address\\
% {\tt\small firstauthor@i1.org}
% % For a paper whose authors are all at the same institution,
% % omit the following lines up until the closing ``}''.
% % Additional authors and addresses can be added with ``\and'',
% % just like the second author.
% % To save space, use either the email address or home page, not both
% \and
% Second Author\\
% Institution2\\
% First line of institution2 address\\
% {\tt\small secondauthor@i2.org}
% }

\maketitle

%%%%%%%%% ABSTRACT
\begin{abstract}
In this study, we propose a self-supervised video denoising method called \qq{restore-from-restored.} This method fine-tunes a pre-trained network by using a pseudo clean video during the test phase. The pseudo clean video is obtained by applying a noisy video to the baseline network.
By adopting a fully convolutional neural network (FCN) as the baseline, we can improve video denoising performance without accurate optical flow estimation and registration steps, in contrast to many conventional video restoration methods, due to the translation equivariant property of the FCN.
Specifically, the proposed method can take advantage of plentiful similar patches existing across multiple consecutive frames (i.e., patch-recurrence); these patches can boost the performance of the baseline network by a large margin.
We analyze the restoration performance of the fine-tuned video denoising networks with the proposed self-supervision-based learning algorithm, and demonstrate that the FCN can utilize recurring patches without requiring accurate registration among adjacent frames. In our experiments, we apply the proposed method to state-of-the-art denoisers and show that our fine-tuned networks achieve a considerable improvement in denoising performance \footnote{Code is available at \url{https://github.com/shlee0/RFR-video-denoising}}.
\end{abstract}

%%%%%%%%% BODY TEXT
\section{Introduction}

% Video restoration is one of the oldest research fields of video processing, which aims to recover the high-quality video frames from the low-quality video. This degradation operation can be formulated with a degradation function $H$ as
% \begin{equation}
% \textbf{Y} = \textbf{H} (\textbf{X}),
% \label{equ_general}
% \end{equation}
% where $\textbf{Y}$, and $\textbf{X}$ are an observed input and a desired clean images. 
% In the case of denoising, $\textbf{H}(\textbf{X})$ add a random noise to the function input image as
% \begin{equation}
% \textbf{H}(\textbf{X}) = \textbf{X} + \mathbf{n},
% \end{equation}
% where $\mathbf{n}$ denotes the noise (\eg, Additive White Gaussian Noise).

Video restoration, which aims to recover the high-quality video frames from the low-quality video, is one of the oldest research fields in video processing.
\if 0
In particular, video denoising which removes noise in the input video frames has been researched considerably, and can be formulated as
\begin{equation}
\textbf{Y} = \textbf{X} + \mathbf{N},
\label{equ_general}
\end{equation}
where $\textbf{Y}$, $\textbf{X}$ and $\mathbf{N}$ denote an observed input frame, desired clean images, and random noise (\eg, Additive White Gaussian Noise) respectively.

Estimating $\textbf{X}$ from $\textbf{Y}$ is a well-known inverse problem. 
\fi
Video denoising, which removes noise in the video frames, has been investigated considerably.
However, estimating a clean image from a corrupted frame is a well-known inverse problem.
To solve such an ill-posed problem, various types of approaches, including prior model, likelihood model, optimization, and deep learning techniques, have been introduced.

\if 0
Video restoration (\eg, SR, denoising, and De-JPEG) is one of the oldest research fields of video processing, whose aim is to recover the high-quality video frames from the video damaged by certain reasons such as noise and compression. This damaging operation is formulated with the degradation function $H$ as
\begin{equation}
\textbf{Y}_{t} = \textbf{H} (\textbf{X}_{t}),
\label{equ_general}
\end{equation}
where $\textbf{Y}_{t}$, and $\textbf{X}_{t}$ are an observed input and a desired clean image at time $t$. It is an inverse problem that predicts $\textbf{X}_{t}$ from given $\textbf{Y}_{t}$ condition on $\textbf{H}$.  In order to solve this inverse problem, based on the natural image prior,  various types of approaches have been introduced including optimization, MAP, sparse coding and deep learning. \cho{If it is possible, it would be good to references about each category of approach types.}
\fi

A common natural image property used for image restoration is patch-recurrence in which similar patches exist within a single image. 
Particularly, patch-recurrence has been considerably studied in single-image super-resolution (SR) methods~\cite{glasner,selfex,SRHRF+}.
%In a video, not only similar patches but also those that are exactly the same can exist over multiple video frames. 
Although these patches can be deformed by camera and/or object motion in a video, patch-recurrence over neighboring video frames is much richer than that of a single image and can further improve the quality of the restored frames~\cite{mahmoudi2005fast,maggioni2012video}.
Moreover, rich patch-recurrence information can greatly help in the fine-tuning of video restoration networks during the test stage, without using ground-truth clean images.% that correspond to the degraded input video frames.%Note that, the ground-truth clean images are not available at the test stage in real-world scenario.

Lehtinen~\etal~\cite{Noise2noise} proposed single-image denoising method (noise-to-noise), which allows the training of the restoration network without ground-truth clean images.
Ehret~\etal~\cite{Ehret} proposed a frame-to-frame training technique, which extends the noise-to-noise training algorithm for video restoration; the frame-to-frame training algorithm can also perform fine-tuning without using the ground-truth clean video by aligning noisy patches among consecutive frames using optical flow.
However, estimating accurate optical flow under large displacements, occlusion, and severe degradation (\eg, noise and blur) is a challenging task. Thus, in this work, we propose a new training algorithm called \qq{restore-from-restored,} which allows fine-tuning of pre-trained networks without using the ground-truth clean video and accurate optical flow for registration.

Our proposed method updates the parameters of pre-trained networks using pseudo clean images, which are outputs of the pre-trained baseline networks from noisy input frames. Our algorithm is simple yet effective for removing noise in video frames and works particularly well with the existence of numerous recurring patches.
That is, we generate pairs of training images, which are composed of the pseudo clean video and its noisy versions, to fine-tune the network.
%In this way, similar patches in different frames are automatically paired with different pseudo patches, and thus an optimal latent patch becomes an average version of these pseudo clean patches.
In practice, pixel locations of the same patches in different video frames vary due to motions, but with the aid of the translation equivariant property of a fully convolutional network (FCN), our algorithm can update the network parameters without using optical flow to align the translated patches only if they are fully convolutional.
We demonstrate the superiority of the proposed algorithm by applying it to state-of-the-art video denoising networks and providing improved denoising results. The contributions of this study are summarized as follows:
\begin{itemize}
\item We propose a novel self-supervised training algorithm to fine-tune fully pre-trained networks without using the clean ground-truth video.
\item We explain why and how the proposed training scheme works with the patch-recurrence property.
\item The proposed method can be easily integrated with state-of-the-art denoising networks and yields state-of-the-art denoising results on the benchmark datasets including not only synthetic but also real noise.
\end{itemize}

%------------------------------------------------------------------------
\section{Related Works}
In this section, we provide a brief overview of recent works that are related to the proposed restoration algorithm, in terms of training with and without using ground-truth clean data.

\noindent\textbf{Training with ground-truth clean data.}
When a set of high-quality images is available, we can generate synthetic degraded images, and train deep neural networks with these images and restore them to their original high-quality state. 
%This paradigm is called \qq{damage and restore}, which is a type of self-supervised learning that most image restoration methods follow.

In the case of image denoising, Xie~\etal~\cite{Denoising12} applied deep neural networks to model the mapping of clean images from noisy input images. They generated pairs of noisy and clean images to train the neural networks.
Since then, numerous studies on the image denoising task have been conducted using deep CNN with train pairs of clean and synthetically noisy images~\cite{DnCNN,IrCNN,FFDNet,RDN,Nonlocal_color,Nonlocal_recurrent,Nonlocal_residual,CBDNet,RIDNet}.
%As the inputs and outputs of networks for image restoration share almost the same information, especially in terms of the low-frequency components, several studies adopted residual learning scheme~\cite{DnCNN,RDN,vdsr}.
Several studies have adopted residual learning schemes to allow deep neural networks and extend the receptive field~\cite{DnCNN,RDN}. Moreover, to incorporate long-range dependencies among pixels, several studies utilized non-local networks~\cite{Nonlocal_residual,Nonlocal_recurrent}. Recent efforts have attempted to deal with unknown noise in real photographs (blind restoration). Guo~\etal~\cite{CBDNet} proposed a two-stage method that consists of noise estimation and non-blind denoising steps. Gao and Grauman~\cite{ondemand} proposed an on-demand learning method to handle various corruption levels for each restoration task including denoising, inpainting, and deblurring. These research trends have also been applied to video restoration problems. For instance, Davy~\etal~\cite{vnlnet} not only incorporated non-local information with a non-local patch search module but also adopted a residual learning scheme for video denoising. FastDVDnet~\cite{fastdvdnet} proposed a cascaded two-step architecture with a modified multi-scale U-Net. FastDVDnet exhibits fast runtimes by avoiding costly explicit motion compensation and handling motion implicitly due to the attributes of its dedicated architecture. Yue~\etal~\cite{rvidenet} proposed RViDeNet to restore real noisy video frames in raw image spaces and achieved state-of-the-art denoising performance. RViDeNet adopts deformable convolution~\cite{DeformConv} to align consecutive frames and utilizes spatio-temporal fusion stages to reconstruct the result.
%and utilizes temporal fusion for aligned features and go through spatial fusion to reconstruct the result.
Moreover, Yue~\etal provided a new video denoising dataset, namely, Captured Raw Video Dataset (CRVD).
%Well-known issues in video restoration are fully utilizing temporal information from multiple frames~\cite{edvr} and predict temporally consistent results~\cite{temporal_consistent}.

Regardless of the techniques used in previous works (\eg, residual learning, non-local network, and model-blind approaches), these studies require the generation of synthetic images for network training. Thus, these supervised approaches hardly deal with datasets where clean images can be rarely obtained (\eg, medical imaging system).
%In addition, they cannot fine-tune the networks to fit the specific input during the test phase.

\noindent\textbf{Training without ground-truth clean data.}
Several attempts have been made recently to learn restoration networks without using ground-truth clean data. Lehtinen~\etal~\cite{Noise2noise} trained a network with pairs of noisy patches under the assumption that the average of many differently corrupted pixels is close to clean data. Then, Krull~\etal~\cite{Noise2void} and Baston and Royer~\cite{Noise2self} introduced self-supervised single image denoising methods without relying on clean data.
%As these methods allow networks to learn without the clean data, pre-trained networks can be fine-tuned given specific input images or videos in the test phase. %In~\cite{zssr}, a method for training SR networks using test input images was presented. This method can utilize the power of deep learning and the information from input images at test time.
Ehret~\etal~\cite{Ehret} introduced a frame-to-frame training method to learn video restoration networks without clean images by extending the strategy proposed in~\cite{Noise2noise} to videos. For removing noise in a certain patch, their method searches corresponding patches among adjacent frames by using optical flow and then warps the patches to create pairs of aligned noisy patches for network training. Frame-to-frame training enables the exploitation of patch-recurrence property within input video frames and fine-tune the pre-trained networks using test inputs. This approach can boost the performance of the existing pre-trained networks because networks can be further optimized without the ground-truth targets at the test stage.

However, one disadvantage of this method~\cite{Ehret} is that accurate optical flow, which is difficult to estimate under large displacements, occlusions and serious damages, is required to fine-tune network. In this study, we overcome the limitations of~\cite{Ehret} by using a new training scheme called \qq{restore-from-restored.} Technically, pseudo clean images are generated from a pre-trained video restoration network and then used as train targets for fine-tuning during the test phase.
It produces a synergy effect with the patch-recurrence property that appears repeatedly over consecutive video frames.
In the following sections, we provide detailed analysis on the proposed method and show the proposed method can boost the performance of the fully pre-trained denoising networks with the help of the patch-recurrence property.

To the best of our knowledge, the proposed method is the first neural approach to boost the performance of pre-trained convolutional video restoration networks without using accurate registration or non-local operation while using recurring patches in the test-phase.

%-------------------------------------------------------------------------
\section{Self-Supervised Video Restoration}

Patch-recurrence within the same scale is rich in natural images~\cite{zssr,singan} and becomes more redundant when multiple neighboring video frames are available~\cite{stsr}.
To utilize this space-time recurring information among given video frames, conventional restoration methods require accurate correspondences between adjacent frames and thus need to compute the optical flow to align the neighboring frames to the reference frames~\cite{vespcn,frvsr,sttn}.

In this work, we present a novel yet simple training algorithm (test-time fine-tuning) that can be applied to video restoration networks. Our fine-tuning algorithm is based on self-supervision and does not require ground-truth clean images.
Moreover, the proposed algorithm allows restoration networks to exploit patch-recurrence without accurate optical flow estimation and registration steps while improving performance by a large margin.
Many convolutional video restoration networks, including state-of-the-art methods, can be easily fine-tuned using our self-supervised training algorithm without changing their original network architecture if they are fully convolutional.

\subsection{Restore-from-restored}

In this section, we explain how we can fine-tune and improve the performance of the pre-trained video restoration networks without using the clean video frames during the test stage.

In general, conventional video restoration networks are trained with labeled ground-truth clean images; these networks learn a function $f_{\theta}$, which maps a corrupted input frame $\textbf{Y}$ to a clean target frame $\textbf{X}$, where $\theta$ denotes the function parameters. Specifically, the network parameter $\theta$ is trained by minimizing the loss function $\textit{L}$ between the network outcome and the training target as
\begin{equation}
Loss(\theta)=\textit{L}(f_{\theta}(\textbf{Y}), \textbf{X}).
\label{equ_general_loss}
\end{equation}
For the loss function $\textit{L}$, common choices are L1 and L2 losses in many denoising approaches~\cite{Noise2self,Noise2noise}. Although image restoration networks trained by minimizing the distance between the network output and the training target can produce highly satisfactory results, these networks can be further upgraded by utilizing redundant spatio-temporal information (\eg, patch-recurrence) over neighboring video frames~\cite{sttn,vnlnet}. However, optical flow estimation networks or non-local operation modules to exploit the recurring patches among different frames are expensive and require additional resources to extract the temporal information~\cite{vespcn,vnlnet,frvsr,Nonlocal_recurrent,Nonlocal_residual}. 

Therefore, we develop a simple and effective fine-tuning algorithm that can exploit patch-recurrence in space-time without explicitly searching similar patches/features through optical flow estimation or non-local operation. To achieve this goal, we assume that we have a fully pre-trained network $f_{\theta_0}$ and use initially denoised video frames $\{\Tilde{\textbf{X}}_1,..., \Tilde{\textbf{X}}_T\}$ as train targets for the fine-tuning, where $\tilde{\mathbf{X}}_t = f_{\theta_0}(\textbf{Y}_t)$ and $t$ denotes the frame index. Although these denoised images are not clean ground-truth images and may include some artifacts, they can be used as pseudo clean targets to fine-tune the network parameter in our study.
Using the pseudo clean images, we can synthesize pseudo noisy images by adding random noise $\textbf{N}$ to the pseudo clean images, and the pairs of pseudo clean and pseudo noisy images can be used to fine-tune the denoising networks by minimizing the loss as follows:
\begin{equation}
Loss(\theta) = \sum_{t=1}^T \textit{L}(f_{\theta}(\Tilde{\textbf{X}}_t+\mathbf{N}),\Tilde{\textbf{X}}_t),
\label{equ_proposed_loss}
\end{equation}
Notably, the ground-truth frames and motion flows are not used during our fine-tuning process. Nevertheless, under an assumption that the distributions of the original noisy input and the corresponding pseudo noisy images are similar, we can update the network parameter $\theta$ by minimizing the proposed loss with the initially denoised frames and their synthetically corrupted counterparts if the networks are fully convolutional.
Assume that we have self-similar noisy patches $\mathbf{y}_a$ and $\mathbf{y}_b$ where $\mathbf{y}_b$ is a translated version of $\mathbf{y}_a$ by a translational motion $A$ (\ie, $A(\mathbf{y}_a)=\mathbf{y}_b$).
Then, we can generate pseudo clean patches $\tilde{\mathbf{x}}_a$ and $\tilde{\mathbf{x}}_b$ with a pre-trained network, and we see that $A(\tilde{\mathbf{x}}_a)=\tilde{\mathbf{x}}_b$ when the network is an FCN, and it yields,
\begin{equation}
\begin{split}
\textit{L}(f_{\theta}(\Tilde{\textbf{x}}_a+\mathbf{N}), \Tilde{\textbf{x}}_a)
&\approx \textit{L}(f_{\theta}(A(\Tilde{\textbf{x}}_a+\mathbf{N})), A(\Tilde{\textbf{x}}_a)) \\
&\approx  \textit{L}(f_{\theta}(\Tilde{\textbf{x}}_b+\mathbf{N}), \Tilde{\textbf{x}}_b).
\end{split}
\end{equation}
Therefore, overall loss in (\ref{equ_proposed_loss}) does not depend on locations of self-similar patches over multiple frames during the back-propagation with FCNs, and our fine-tuned network predicts the averaged version of denoised self-similar patches with L2 loss when one of the self-similar noisy patches is given as input.
%If random translational motion $w$ is applied to each frame, we obtain translated frames, $\tilde{\mathbf{x}}_t^{w}$.
%Using the loss function in (\ref{equ_proposed_loss}), we optimize the network parameter by minimizing 
%$\sum_{t=1}^T\textit{L}(f_{\theta}(\Tilde{\textbf{x}}_t^w+\mathbf{N}), \Tilde{\textbf{x}}_t^w)$
We call this process \qq{restore-from-restored} training, since the proposed method is relying on the initially restored (denoised) frames. We repeat our training algorithm several times until convergence and achieve considerable improvement over the fully pre-trained initial network.

\subsection{Frame-to-frame vs. Restore-from-restored}
The notion of our \qq{restore-from-restored} algorithm is based on recent noise-to-noise training mechanism by Lehtinen~\etal~\cite{Noise2noise}.
Noise-to-noise demonstrated that image restoration networks can be trained without using ground-truth clean data for certain types of noise (\eg, zero-mean noise, such as Gaussian noise and Bernoulli noise) and can be extended into the frame-to-frame approach by Ehret~\etal~\cite{Ehret} to process a video.

\paragraph{Exploiting space-time patch-recurrence.}
% \noindent\textbf{Exploiting space-time patch-recurrence.}
The frame-to-frame training algorithm~\cite{Ehret} allows the network to train with self-supervision during the test-stage. Specifically, the network is fine-tuned with two aligned noisy frames by minimizing the loss as
\begin{equation}
Loss(\theta)=\sum_{t=1}^T\textit{L}(f_{\theta}(\textbf{Y}_t), \textbf{Y}^w_{t-1}),
\label{equ_f2f_loss}
\end{equation}
where $\textbf{Y}^w_{t-1}$ denotes the warped version of the noisy frame $\textbf{Y}_{t-1}$ and is aligned to the reference frame $\textbf{Y}_t$. Thus, the calculation of optical flow for registration is necessary in \eqref{equ_f2f_loss}. However, the accurate estimation of optical flow is difficult to achieve in some conditions, such as severe degradation, large displacement between frames.

By contrast, our proposed loss in \eqref{equ_proposed_loss} does not need warping and alignment. If the denoising network $f_{\theta}$ is an FCN, then our method can maintain the performance without accurate registration due to the translation equivariant nature of FCN~\cite{cohen2016group}. That is, our \qq{restore-from-restored} approach is not disturbed by the existence of large translational motions compared with optical flow-based methods.

\paragraph{Noise reduction.}
% \noindent\textbf{Noise reduction.}
Assume that a set of perfectly aligned images $\{\textbf{Y}_1,...,\textbf{Y}_T\}$ (\eg, burst mode images from a camera on a tripod) is given, and these images are corrupted by zero-mean Gaussian random noise whose standard deviation is $\sigma$. 
Using the frame-to-frame training algorithm~\cite{Ehret}, the denoised frames become an averaged version of noisy inputs (=$\frac{1}{T}\sum_{t=1}^T \textbf{Y}_t$) with the fine-tuned parameter, and the noise variance of the denoised frame is reduced to $\frac{1}{T}{\sigma}^2$ (refer to the Appendix in~\cite{Noise2noise} for details). 

By contrast, the latent frame predicted from a fine-tuned parameter by minimizing the L2 version of the proposed loss in \eqref{equ_proposed_loss} is $\frac{1}{T}\sum_{t=1}^T \tilde{\mathbf{X}}_t$. Thus, the noise variance of our denoising result becomes $\frac{1}{T}{\sigma_{\theta_0}}^2$, where $\sigma_{\theta_0}$ denotes the mean standard deviation of the remaining noise in $\tilde{\mathbf{X}}_t$. In general, as $f_{\theta_0}$ is a fully pre-trained network, the noise level of the residual noise $\sigma_{\theta_0}$ is much lower than the original noise level $\sigma$ (\ie, $\sigma_{\theta_0}<<\sigma$). Therefore, the noise variance of the latent frame from our \qq{restore-from-restored} algorithm is much lower than that from the frame-to-frame training.
Notably, our algorithm can achieve better results when there is a rich patch-recurrence, and our algorithm can show space-time varying denoising performance. For example, noisy regions where structures are highly repeating can be restored much better than regions with slightly repeating or unique patterns.

%-------------------------------------------------------------------------
\section{Proposed Method}

\begin{algorithm}[h]
\setstretch{1.2}

\textbf{Input:} degraded video frames $\{\textbf{Y}_1,...,\textbf{Y}_T\}$ 

\textbf{Output:} denoised video frames $\tilde{{\textbf{X}}}^K$ \\

\textbf{Require:} pre-trained network $f_{\theta_0}$, iteration number K, learning rate $\alpha$\\

\nl i $\leftarrow$ 0

\While{i $\leq$ K}
{
    \nl \ForEach{t}{
    \nl Restore: $\tilde{\textbf{X}}_t^i \leftarrow f_{\theta_i}(\textbf{Y}_t)$
    }
    
    \nl $\tilde{{\textbf{X}}}^i \leftarrow \{{\tilde{\textbf{X}}_1^i,...,\tilde{\textbf{X}}_T^i}\}$

    \nl $\textit{Loss}(\theta_i) =
    \sum_{t=1}^T \textit{L}(f_{\theta_i}(\Tilde{\textbf{X}}_t^i+\textbf{N}), \Tilde{\textbf{X}}_t^i) +
    \sum_{t=1}^T\textit{L}(f_{\theta_i}(\Tilde{\textbf{X}}_t^0+\textbf{N}), \Tilde{\textbf{X}}_t^0)$

    \nl $\theta_{i+1} \leftarrow \theta_i - \alpha \nabla_{\theta_i}  \textit{Loss}(\theta_i)$
    
    \nl i $\leftarrow $ i+1  
}

\nl \textbf{Return:} $\tilde{{\textbf{X}}}^K$

\caption{Offline video denoising algorithm}

\label{algorithm_offline}

\end{algorithm}

\begin{algorithm}
\setstretch{1.2}
\textbf{Input:} degraded frame $\textbf{Y}_t$ at time-step $t$

\textbf{Output:} restored frame $\tilde{\textbf{X}}_{t}$, fine-tuned network $f_{\theta_t}$

\textbf{Require:}  fine-tuned network $f_{\theta_{t-1}}$,  restored frame at the previous time-step $\tilde{\textbf{X}}_{t-1}$, learning rate $\alpha$

\nl $\textit{Loss}(\theta_{t-1}) = \textit{L} (f_{\theta_{t-1}}(\Tilde{\textbf{X}}_{t-1}+\textbf{N}), \Tilde{\textbf{X}}_{t-1})$

\nl $\theta_t \leftarrow \theta_{t-1} - \alpha \nabla_{\theta_{t-1}}  \textit{Loss}(\theta_{t-1})$

\nl Restore: $\Tilde{\textbf{X}}_t \leftarrow f_{\theta_t}(\textbf{Y}_t)$

\textbf{Return:} $\Tilde{\textbf{X}}_t, f_{\theta_t}$  \tcp{Will be reused in the next time step}

\caption{Online video denoising algorithm}
\label{algorithm_online}
\end{algorithm}

By minimizing the proposed loss function in~\eqref{equ_proposed_loss}, we can restore clean images from degraded ones in offline and online manners, similar to~\cite{Ehret}.
In the offline video denoising mode, we can take advantage of all the given frames for fine-tuning the network parameter. In the online denoising mode, fine-tuning and denoising are conducted successively in a sequential fashion (frame-by-frame).
%Specifically, we provide temporally varying network parameters; thus, each frame is denoised with a different network parameter.

\paragraph{Offline denoising.}
% \noindent\textbf{Offline denoising.}
The offline video denoising algorithm is elaborated in \algoref{algorithm_offline}.
For offline fine-tuning at each train step $i$, we remove the noise in the input noisy images $\{\textbf{Y}_1,...,\textbf{Y}_T\}$ using denoising network $f_{\theta_i}$ and obtain denoised images $\{\Tilde{\textbf{X}}_1^i,..., \Tilde{\textbf{X}}_T^i\}$. Next, we use pairs of training images $\{(\Tilde{\textbf{X}}_1^i+\mathbf{N}, \Tilde{\textbf{X}}_1^i),..., (\Tilde{\textbf{X}}_T^i+\mathbf{N}, \Tilde{\textbf{X}}_T^i)\}$ to fine-tune the network parameter by minimizing the loss function in \eqref{equ_proposed_loss}. 
To avoid over-fitting and generating over-smoothed results, we use the initially denoised frames $\{\Tilde{\textbf{X}}_1^0,..., \Tilde{\textbf{X}}_T^0\}$ as additional train targets during the fine-tuning stages. In our supplementary material, we demonstrate the effect of the second loss term in~\algoref{algorithm_offline} by comparing results with and without using the initial frames during offline fine-tuning when $K$ is large.

\paragraph{Online denoising.}
% \noindent\textbf{Online denoising.}
We can also adapt the network parameter in a sequential manner. Our online video denoising algorithm is given in \algoref{algorithm_online}. In contrast to the offline denoising mode, our online denoising algorithm uses only a previous frame for the update at each time step. To do so, we slightly modify the proposed loss function in~\eqref{equ_proposed_loss} for the time step $t$, to take a single pair of images $(\Tilde{\textbf{X}}_{t-1}+\textbf{N},\tilde{\textbf{X}}_{t-1})$ which is previous pseudo clean and noisy frames. In~\algoref{algorithm_online}, temporally varying parameter $\theta_{t}$ denotes the fine-tuned network parameter at time $t$; thus, each frame is denoised with a different network parameter.
%Therefore, the network parameter becomes a time-varying variable in our online denoising algorithm.

%-------------------------------------------------------------------------
\section{Experiments}
In our experiments, we apply our offline and online denoising algorithms to conventional denoising networks, including state-of-the-art video denoising methods, and evaluate the denoising performance quantitatively and qualitatively.
Please refer to our supplementary material for additional experimental results, including video clips.

\subsection{Implementation Details}
We use officially available fully pre-trained network parameters for our baseline denoisers.
During the fine-tuning period, we minimize the L2 loss for the proposed offline and online denoising algorithms.
%to update network parameters as it slightly outperforms L1 loss empirically in our experiments. 
We also use Adam for the updates, with a learning rate of 1e-5.
We measure the performance in terms of PSNR and SSIM on RGB color channels for the objective evaluation.

First, for Gaussian noise removal, we use VNLnet~\cite{vnlnet} and FastDVDnet~\cite{fastdvdnet}.
% For denoising, we use FCN-based DnCNN~\cite{DnCNN} and VNLnet~\cite{vnlnet} to demonstrate the performance of the proposed methods. DnCNN is a typical convolutional networks and used as backbone or baseline model in many other works including~\cite{Noise2void,Noise2self,Ehret}.
VNLnet is a state-of-the art video denoising method and shows the best denoising performance.
Although VNLnet is integrated with the non-local operation module, we show that our algorithm can be easily applied to networks with the non-local module and can further improve the performance of the baseline networks.
For the evaluation, we use 7 video clips consisting of 100 sequences each on the Derf database \footnote{\textit{https://media.xiph.org/video/derf/}} and additional 7 video clips in the DAVIS video segmentation challenge dataset~\cite{davis}.
We use down-scaled video frames (960$\times$540), and generate noisy input videos by adding Gaussian random noise with different noise levels ($\sigma$=15, 25, 40).

Next, for real noise removal, we use RViDeNet~\cite{rvidenet} and evaluate on the CRVD dataset~\cite{rvidenet}. We improve the baseline performance with the proposed fine-tuning algorithm in the sRGB space. To handle real noise, we synthesize random noise $\textbf{N}$ with the noise model used in the pre-training process of RViDeNet. Refer to our supplementary material for detailed settings.

\subsection{Denoising Performance}

\begin{figure}[]
\centering
\includegraphics[width=1.0\linewidth]{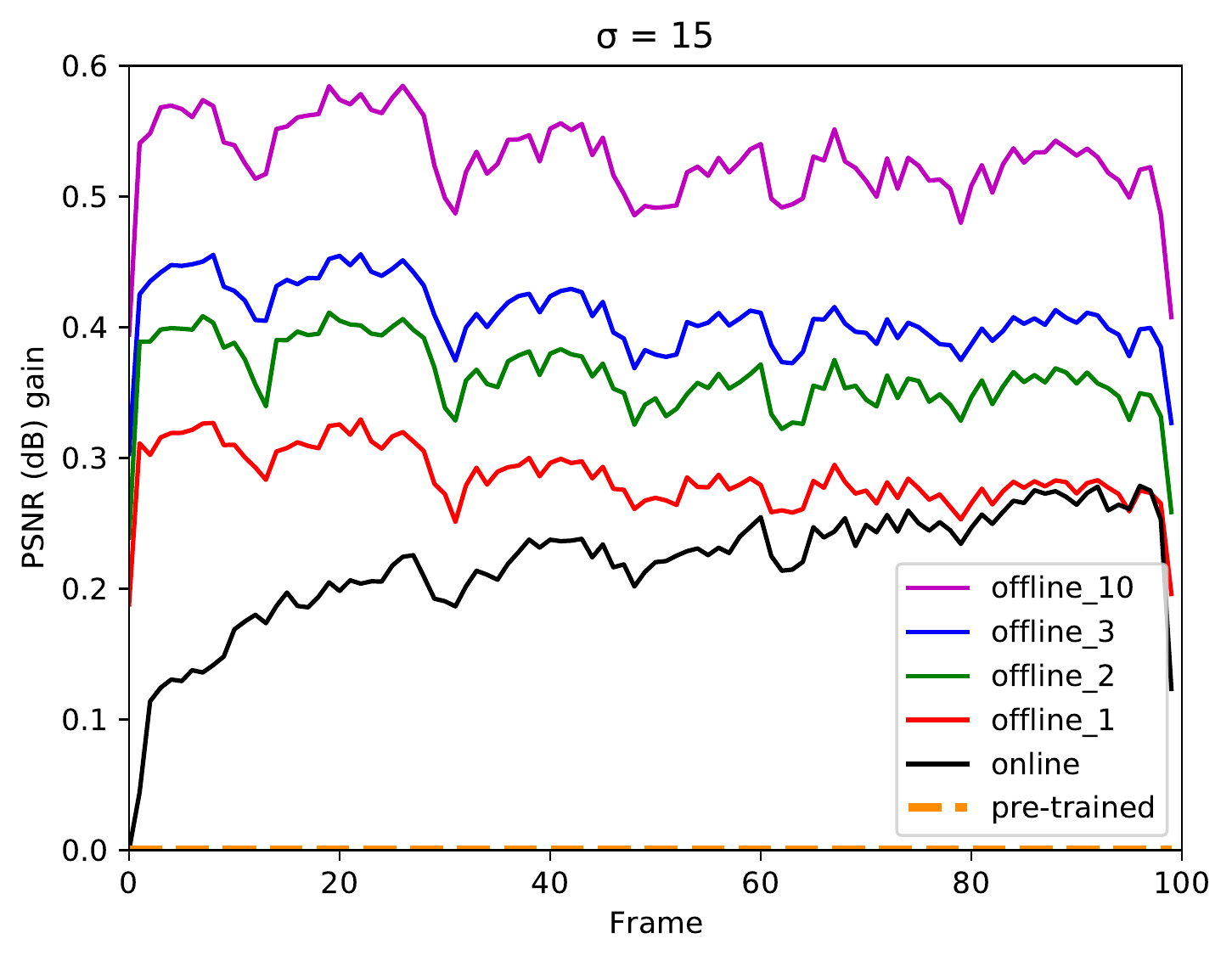}
\caption{Performance gains from our online and offline denoising algorithms. Baseline network is FastDVDnet~\cite{fastdvdnet}. Differences of PSNR values before and after fine-tuning are measured on Derf datasets on $\sigma$ = 15. Number $i$ in \qq{offline\_$i$} denotes the number of steps (\ie, $K$). Note that the rapid rise and drop of performance at the very first and last time step are due to a usability of adjacent frames.}
\label{fig_psnrgain}
\end{figure}

\begin{figure*}[]
\centering
\includegraphics[width=1.0\linewidth]{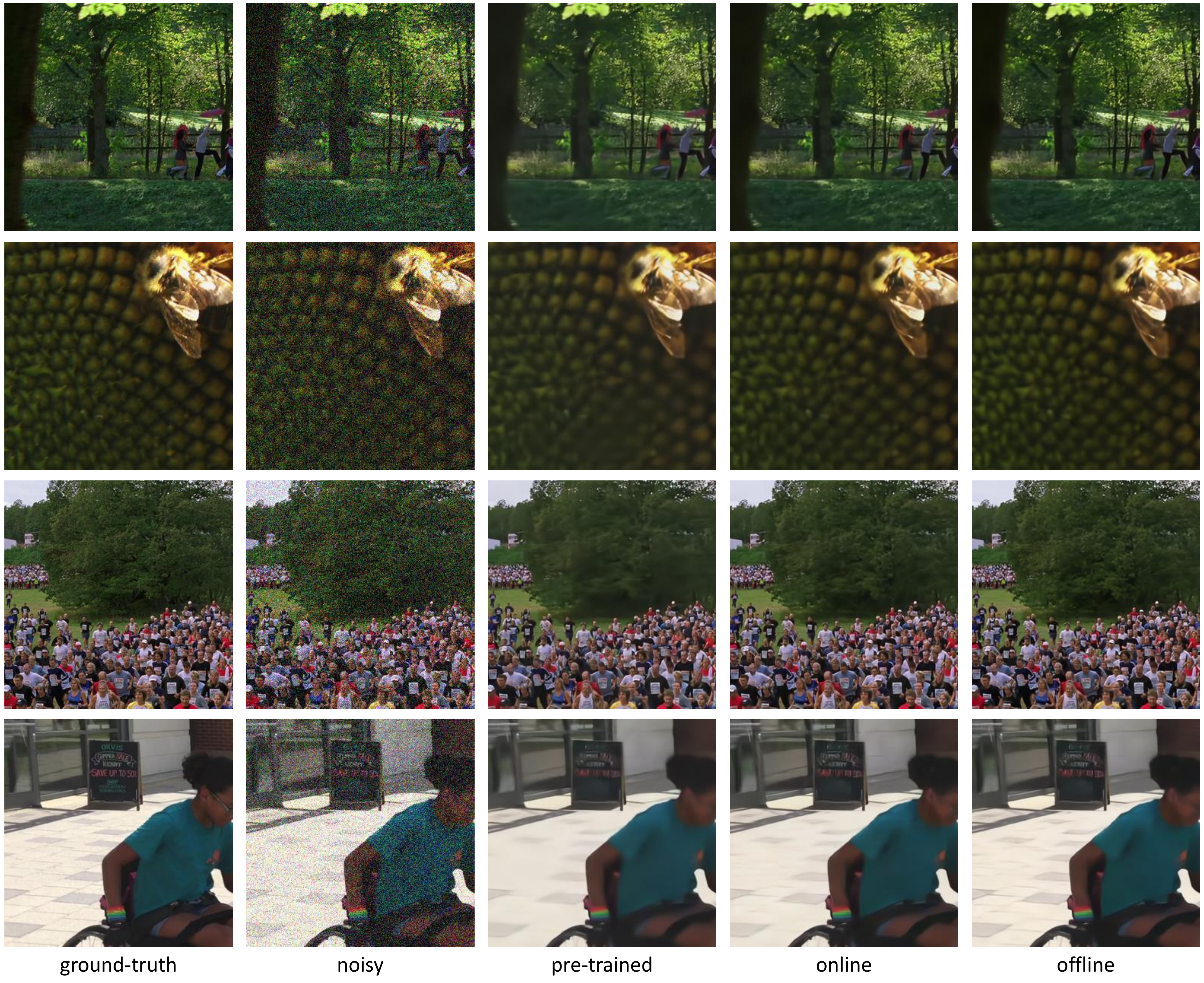}
%\vspace{-2Ex}
\caption{Denoising results with FastDVDnet~\cite{fastdvdnet} on Derf testsets corrupted by Gaussian noise ($\sigma$ = 40). Visual comparisons with our online and offline ($K$ = 10 in~\algoref{algorithm_offline}) update procedures.} 
\label{visualresult_gaussian}
%\vspace{-3Ex}
\end{figure*}

\begin{figure}[]
\centering
\includegraphics[width=1.0\linewidth]{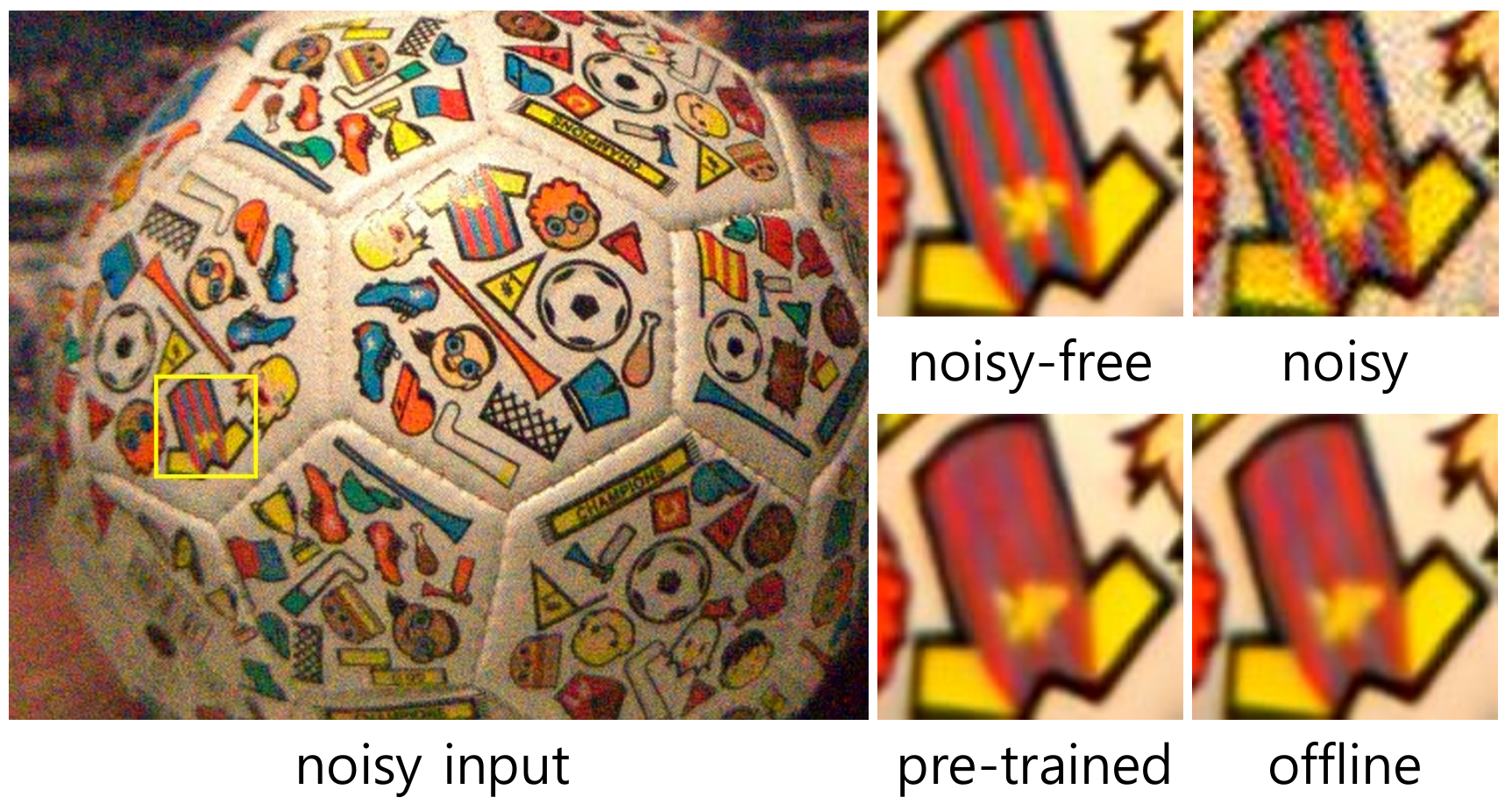}
%\vspace{-2Ex}
\caption{Denoising results with RViDeNet on the CRVD~\cite{rvidenet} dataset which includes real noise. We use $K$ = 10, for our offline denoising algorithm.} 
\label{visualresult_real}
\end{figure}

\paragraph{Quantitative results.}
First, we fine-tune FastDVDnet~\cite{fastdvdnet} on the Derf dataset with the proposed offline and online denoising algorithms to evaluate the Gaussian denoising performance; the performance gains are depicted in~\figref{fig_psnrgain}.
We achieve consistently improved denoising performance over the baseline as the number of iterations (\ie, $K$) increases with the offline denoising algorithm. Meanwhile, we obtain improved results as the frame number increases with our online denoising algorithm because the proposed online method enables sequential parameter update.
In~\tabref{denosing_result_derf} and~\tabref{denosing_result_davis}, we provide the PSNR and SSIM values of the denoising results from our algorithms on the Derf and DAVIS testsets. These results show that the online and offline denoising algorithms can produce steadily better results than the baseline models, and the offline denoising algorithm with 10 updates (\ie, $K$ = 10) achieves the best performance. In~\tabref{tab:result_rvidenet}, we add quantitative denoising results on the real noise dataset, CRVD~\cite{rvidenet}. As CRVD includes only short sequences ($T$ = 7), we only evaluate the performance of the offline restoration algorithm; the results show that the proposed learning algorithm can effectively remove real noise remarkably and enhance the denoising results.

% \paragraph{Visual results.}
\noindent\textbf{Visual results.}
In~\figref{visualresult_gaussian}, we provide qualitative comparison results. The input images are corrupted with high-level Gaussian noise ($\sigma$ = 40), and FastDVDnet is fine-tuned by our offline and online restoration algorithms. We also provide real noise denoising results with fine-tuned RViDeNet in~\figref{visualresult_real}. Our methods can produce much better visual results and restore tiny details compared with the initially pre-trained network.

% \paragraph{Run-time.}
\noindent\textbf{Run-time.}
We report the run-time for a single update step on the NVIDIA Tesla V100 graphics unit. FastDVDnet and VNLnet take approximately 0.16 and 0.4 second to handle a 960$\times$540 input frame with our online denoising algorithm in~\algoref{algorithm_online}; RViDeNet takes approximately 3.0 second to process the 256$\times$256 input frame with our offline denoising method in~\algoref{algorithm_offline}.

\begin{table*}[]
\centering
\resizebox{\textwidth}{!}{%
\begin{tabular}{|c|c|c|c|c|c|c|c|c|c|}
\hline
Method                  & $\sigma$ & crowd                                                         & park joy                                                      & pedestrian                                                    & station                                                       & sunflower                                                     & touchdown                                                     & tractor                                                       & Average                                                       \\ \hline \hline
\multirow{3}{*}{FastDVDnet}  & 15    & \begin{tabular}[c]{@{}c@{}}31.17/0.9208\\ 31.32/0.9233\\ 31.57/0.9263\end{tabular} & \begin{tabular}[c]{@{}c@{}}30.46/0.8959\\ 30.72/0.9058\\ 30.95/0.9108\end{tabular} & \begin{tabular}[c]{@{}c@{}}38.00/0.9522\\ 38.20/0.9537\\ 38.44/0.9549\end{tabular} & \begin{tabular}[c]{@{}c@{}}36.65/0.9323\\ 36.83/0.9338\\ 37.10/0.9363\end{tabular} & \begin{tabular}[c]{@{}c@{}}37.95/0.9525\\ 38.38/0.9554\\ 39.03/0.9596\end{tabular} & \begin{tabular}[c]{@{}c@{}}35.98/0.9109\\ 36.06/0.9123\\ 36.23/0.9138\end{tabular} & \begin{tabular}[c]{@{}c@{}}34.03/0.9263\\ 34.26/0.9283\\ 34.63/0.9320\end{tabular} & \begin{tabular}[c]{@{}c@{}}34.89/0.9273\\ 35.11/0.9304\\ \textbf{35.42/0.9334}\end{tabular} \\ \cline{2-10} 
                        & 25    & \begin{tabular}[c]{@{}c@{}}29.01/0.8813\\ 29.32/0.8892\\ 29.59/0.8943\end{tabular} & \begin{tabular}[c]{@{}c@{}}28.34/0.8373\\ 28.79/0.8610\\ 29.06/0.8719\end{tabular} & \begin{tabular}[c]{@{}c@{}}35.76/0.9328\\ 36.27/0.9369\\ 36.62/0.9393\end{tabular} & \begin{tabular}[c]{@{}c@{}}34.96/0.9050\\ 35.12/0.9076\\ 35.38/0.9107\end{tabular} & \begin{tabular}[c]{@{}c@{}}34.96/0.9306\\ 36.14/0.9389\\ 37.00/0.9455\end{tabular} & \begin{tabular}[c]{@{}c@{}}33.97/0.8667\\ 34.09/0.8678\\ 34.33/0.8703\end{tabular} & \begin{tabular}[c]{@{}c@{}}31.99/0.8946\\ 32.32/0.8990\\ 32.78/0.9051\end{tabular} & \begin{tabular}[c]{@{}c@{}}32.71/0.8926\\ 33.15/0.9000\\ \textbf{33.54/0.9053}\end{tabular} \\ \cline{2-10} 
                        & 40    & \begin{tabular}[c]{@{}c@{}}26.43/0.8159\\ 27.21/0.8375\\ 27.67/0.8509\end{tabular} & \begin{tabular}[c]{@{}c@{}}25.80/0.7441\\ 26.82/0.7927\\ 27.27/0.8192\end{tabular} & \begin{tabular}[c]{@{}c@{}}32.44/0.9003\\ 33.93/0.9118\\ 34.74/0.9190\end{tabular}  & \begin{tabular}[c]{@{}c@{}}32.89/0.8634\\ 33.34/0.8709\\ 33.42/0.8717\end{tabular} & \begin{tabular}[c]{@{}c@{}}30.66/0.8884\\ 33.39/0.9129\\ 34.93/0.9263\end{tabular} & \begin{tabular}[c]{@{}c@{}}31.78/0.8022\\ 32.18/0.8095\\ 32.56/0.8140\end{tabular} & \begin{tabular}[c]{@{}c@{}}29.69/0.8490\\ 30.32/0.8594\\ 30.98/0.8697\end{tabular} & \begin{tabular}[c]{@{}c@{}}29.95/0.8376\\ 31.03/0.8564\\ \textbf{31.69/0.8679}\end{tabular} \\ \hline
\multirow{3}{*}{VNLnet} & 15    & \begin{tabular}[c]{@{}c@{}}32.68/0.9373\\ 32.83/0.9386\\ 33.00/0.9401\end{tabular} & \begin{tabular}[c]{@{}c@{}}32.19/0.9200\\ 32.47/0.9279\\ 32.72/0.9328\end{tabular} & \begin{tabular}[c]{@{}c@{}}38.85/0.9567\\ 38.94/0.9570\\ 39.05/0.9576\end{tabular}  & \begin{tabular}[c]{@{}c@{}}38.51/0.9501\\ 38.60/0.9508\\ 38.72/0.9515\end{tabular} & \begin{tabular}[c]{@{}c@{}}39.58/0.9628\\ 39.87/0.9640\\ 40.22/0.9661\end{tabular} & \begin{tabular}[c]{@{}c@{}}37.37/0.9347\\ 37.41/0.9348\\ 37.52/0.9350\end{tabular} & \begin{tabular}[c]{@{}c@{}}35.12/0.9378\\ 35.30/0.9393\\ 35.54/0.9416\end{tabular} & \begin{tabular}[c]{@{}c@{}}36.33/0.9428\\ 36.49/0.9446\\ \textbf{36.68/0.9464}\end{tabular} \\ \cline{2-10} 
                        & 25    & \begin{tabular}[c]{@{}c@{}}30.07/0.9013\\ 30.34/0.9063\\ 30.58/0.9105\end{tabular} & \begin{tabular}[c]{@{}c@{}}29.48/0.8651\\ 30.00/0.8849\\ 30.33/0.8970\end{tabular} & \begin{tabular}[c]{@{}c@{}}36.15/0.9353\\ 36.58/0.9375\\ 36.93/0.9397\end{tabular} & \begin{tabular}[c]{@{}c@{}}36.57/0.9263\\ 36.69/0.9280\\ 36.83/0.9293\end{tabular} & \begin{tabular}[c]{@{}c@{}}36.06/0.9430\\ 37.37/0.9494\\ 38.02/0.9539\end{tabular} & \begin{tabular}[c]{@{}c@{}}35.21/0.8976\\ 35.32/0.8991\\ 35.49/0.8994\end{tabular} & \begin{tabular}[c]{@{}c@{}}32.83/0.9072\\ 33.11/0.9106\\ 33.41/0.9142\end{tabular} & \begin{tabular}[c]{@{}c@{}}33.77/0.9108\\ 34.20/0.9165\\ \textbf{34.51/0.9206}\end{tabular} \\ \cline{2-10} 
                        & 40    & \begin{tabular}[c]{@{}c@{}}27.09/0.8366\\ 27.85/0.8536\\ 28.35/0.8648\end{tabular} & \begin{tabular}[c]{@{}c@{}}26.51/0.7724\\ 27.61/0.8153\\ 28.18/0.8425\end{tabular} & \begin{tabular}[c]{@{}c@{}}32.48/0.8992\\ 33.78/0.9078\\ 34.71/0.9154\end{tabular} & \begin{tabular}[c]{@{}c@{}}33.91/0.8817\\ 34.36/0.8879\\ 34.64/0.8910\end{tabular} & \begin{tabular}[c]{@{}c@{}}31.01/0.8975\\ 34.09/0.9231\\ 35.62/0.9343\end{tabular} & \begin{tabular}[c]{@{}c@{}}32.33/0.8122\\ 32.71/0.8199\\ 33.06/0.8220\end{tabular} & \begin{tabular}[c]{@{}c@{}}30.09/0.8569\\ 30.71/0.8672\\ 31.33/0.8760\end{tabular} & \begin{tabular}[c]{@{}c@{}}30.49/0.8509\\ 31.57/0.8678\\ \textbf{32.27/0.8780}\end{tabular} \\ \hline
\end{tabular}%
}
\caption{Denoising results with FastDVDnet~\cite{fastdvdnet} and VNLnet~\cite{vnlnet} on the Derf testset with different Gaussian noise levels ($\sigma$ = 15, 25, 40). For each network architecture and each noise level, the PSNR and SSIM results of the baseline, online learning (\algoref{algorithm_online}) and offline learning (\algoref{algorithm_offline}) are listed in each box from top to bottom. The best average results are written in bold letters. %Average PSNR values by baseline, \algoref{algorithm_online}, and \algoref{algorithm_offline} are given from top to bottom.
}
\label{denosing_result_derf}
\end{table*}

\begin{table*}[]
\centering
\resizebox{\textwidth}{!}{%
\begin{tabular}{|c|c|c|c|c|c|c|c|c|c|}
\hline
Method                  & $\sigma$ & chamaleon                                                     & giant-slalom                                                  & girl-dog                                                      & hoverboard                                                    & monkeys-trees                                                 & salsa                                                         & subway                                                        & Average                                                       \\ \hline \hline
\multirow{3}{*}{FastDVDnet}  & 15    & \begin{tabular}[c]{@{}c@{}}36.65/0.9697\\ 36.73/0.9702\\ 36.92/0.9711\end{tabular} & \begin{tabular}[c]{@{}c@{}}40.79/0.9685\\ 41.08/0.9708\\ 41.29/0.9717\end{tabular} & \begin{tabular}[c]{@{}c@{}}34.25/0.9183\\ 34.29/0.9202\\ 34.35/0.9214\end{tabular} & \begin{tabular}[c]{@{}c@{}}39.55/0.9613\\ 39.63/0.9617\\ 39.71/0.9616\end{tabular} & \begin{tabular}[c]{@{}c@{}}31.59/0.9567\\ 31.63/0.9569\\ 31.75/0.9578\end{tabular} & \begin{tabular}[c]{@{}c@{}}33.41/0.9440\\ 33.62/0.9519\\ 33.84/0.9626\end{tabular} & \begin{tabular}[c]{@{}c@{}}37.56/0.9246\\ 39.20/0.9601\\ 40.03/0.9741\end{tabular} & \begin{tabular}[c]{@{}c@{}}36.26/0.9490\\ 36.60/0.9560\\ \textbf{36.84/0.9601}\end{tabular} \\ \cline{2-10} 
                        & 25    & \begin{tabular}[c]{@{}c@{}}33.98/0.9534\\ 34.22/0.9550\\ 34.53/0.9567\end{tabular} & \begin{tabular}[c]{@{}c@{}}38.67/0.9550\\ 38.94/0.9577\\ 39.16/0.9588\end{tabular} & \begin{tabular}[c]{@{}c@{}}31.61/0.8590\\ 31.73/0.8639\\ 31.88/0.8679\end{tabular} & \begin{tabular}[c]{@{}c@{}}37.36/0.9486\\ 37.54/0.9494\\ 37.82/0.9504\end{tabular} & \begin{tabular}[c]{@{}c@{}}28.71/0.9161\\ 28.74/0.9164\\ 28.84/0.9184\end{tabular} & \begin{tabular}[c]{@{}c@{}}30.17/0.9002\\ 30.59/0.9143\\ 30.93/0.9328\end{tabular} & \begin{tabular}[c]{@{}c@{}}33.48/0.8737\\ 36.32/0.9334\\ 37.95/0.9637\end{tabular} & \begin{tabular}[c]{@{}c@{}}33.43/0.9151\\ 34.01/0.9272\\ \textbf{34.44/0.9355}\end{tabular} \\ \cline{2-10} 
                        & 40    & \begin{tabular}[c]{@{}c@{}}30.97/0.9263\\ 31.58/0.9311\\ 32.30/0.9364\end{tabular} & \begin{tabular}[c]{@{}c@{}}36.53/0.9384\\ 36.84/0.9410\\ 37.16/0.9428\end{tabular} & \begin{tabular}[c]{@{}c@{}}29.01/0.7785\\ 29.44/0.7907\\ 29.84/0.7985\end{tabular} & \begin{tabular}[c]{@{}c@{}}34.82/0.9315\\ 35.32/0.9338\\ 35.92/0.9361\end{tabular} & \begin{tabular}[c]{@{}c@{}}26.29/0.8506\\ 26.32/0.8529\\ 26.38/0.8546\end{tabular} & \begin{tabular}[c]{@{}c@{}}26.76/0.8162\\ 27.65/0.8500\\ 28.26/0.8819\end{tabular} & \begin{tabular}[c]{@{}c@{}}28.83/0.8070\\ 32.88/0.8885\\ 35.85/0.9505\end{tabular} & \begin{tabular}[c]{@{}c@{}}30.46/0.8640\\ 31.43/0.8840\\ \textbf{32.25/0.9001}\end{tabular} \\ \hline
\multirow{3}{*}{VNLnet} & 15    & \begin{tabular}[c]{@{}c@{}}37.30/0.9724\\ 37.37/0.9725\\ 37.50/0.9734\end{tabular} & \begin{tabular}[c]{@{}c@{}}42.31/0.9751\\ 42.32/0.9751\\ 42.40/0.9753\end{tabular} & \begin{tabular}[c]{@{}c@{}}35.68/0.9407\\ 35.73/0.9418\\ 35.81/0.9428\end{tabular} & \begin{tabular}[c]{@{}c@{}}39.83/0.9626\\ 39.82/0.9621\\ 39.88/0.9621\end{tabular} & \begin{tabular}[c]{@{}c@{}}34.87/0.9792\\ 34.94/0.9794\\ 35.02/0.9798\end{tabular} & \begin{tabular}[c]{@{}c@{}}34.04/0.9461\\ 34.21/0.9517\\ 34.39/0.9616\end{tabular} & \begin{tabular}[c]{@{}c@{}}37.60/0.9188\\ 39.42/0.9522\\ 40.41/0.9720\end{tabular} & \begin{tabular}[c]{@{}c@{}}37.37/0.9564\\ 37.69/0.9621\\ \textbf{37.97/0.9667}\end{tabular} \\ \cline{2-10} 
                        & 25    & \begin{tabular}[c]{@{}c@{}}34.46/0.9567\\ 34.66/0.9576\\ 34.98/0.9595\end{tabular} & \begin{tabular}[c]{@{}c@{}}39.75/0.9604\\ 39.81/0.9606\\ 39.91/0.9610\end{tabular} & \begin{tabular}[c]{@{}c@{}}32.76/0.8887\\ 32.93/0.8947\\ 33.09/0.8973\end{tabular} & \begin{tabular}[c]{@{}c@{}}37.58/0.9501\\ 37.66/0.9497\\ 37.89/0.9500\end{tabular} & \begin{tabular}[c]{@{}c@{}}31.96/0.9591\\ 32.02/0.9596\\ 32.09/0.9598\end{tabular} & \begin{tabular}[c]{@{}c@{}}30.60/0.9015\\ 31.03/0.9134\\ 31.40/0.9314\end{tabular} & \begin{tabular}[c]{@{}c@{}}32.88/0.8642\\ 35.66/0.9085\\ 37.98/0.9598\end{tabular} & \begin{tabular}[c]{@{}c@{}}34.28/0.9258\\ 34.83/0.9349\\ \textbf{35.33/0.9456}\end{tabular} \\ \cline{2-10} 
                        & 40    & \begin{tabular}[c]{@{}c@{}}31.26/0.9302\\ 31.81/0.9338\\ 32.57/0.9391\end{tabular} & \begin{tabular}[c]{@{}c@{}}37.23/0.9425\\ 37.41/0.9431\\ 37.65/0.9438\end{tabular} & \begin{tabular}[c]{@{}c@{}}29.71/0.8061\\ 30.24/0.8224\\ 30.71/0.8306\end{tabular} & \begin{tabular}[c]{@{}c@{}}34.96/0.9335\\ 35.31/0.9343\\ 35.93/0.9366\end{tabular} & \begin{tabular}[c]{@{}c@{}}29.13/0.9208\\ 29.23/0.9236\\ 29.30/0.9242\end{tabular} & \begin{tabular}[c]{@{}c@{}}27.05/0.8198\\ 27.99/0.8510\\ 28.69/0.8798\end{tabular} & \begin{tabular}[c]{@{}c@{}}28.55/0.8052\\ 32.05/0.8639\\ 35.65/0.9433\end{tabular} & \begin{tabular}[c]{@{}c@{}}31.13/0.8797\\ 32.01/0.8960\\ \textbf{32.93/0.9139}\end{tabular} \\ \hline
\end{tabular}%
}
\caption{Denoising results with FastDVDnet~\cite{fastdvdnet} and VNLnet~\cite{vnlnet} on the DAVIS testset with different Gaussian noise levels ($\sigma$ = 15, 25, 40). For each network architecture and each noise level, the PSNR and SSIM results of the baseline, online learning (\algoref{algorithm_online}) and offline learning (\algoref{algorithm_offline}) are listed in each box from top to bottom. The best average results are written in bold letters.}
\label{denosing_result_davis}
\end{table*}

\begin{table}[]
\centering
\begin{tabular}{|c|c|c|c|}
\hline
     & Noisy & RViDeNet & Ours   \\ \hline
PSNR & 31.79 & 39.95    & \textbf{40.13}  \\ \hline
SSIM & 0.7517 & 0.9792   & \textbf{0.9795} \\ \hline
\end{tabular}
\caption{Denoising results with RViDeNet on the CRVD~\cite{rvidenet} dataset with real noise. The proposed offline learning method (\algoref{algorithm_offline}) show the quantitatively better results than the baseline (RViDeNet) and the results are written in bold letters.}
\label{tab:result_rvidenet}
\end{table}

\subsection{Comparison with the frame-to-frame}

\begin{figure}[]
\centering
\begin{minipage}[c]{\linewidth}
\centering
    \includegraphics[width=\linewidth, height=0.5\linewidth]{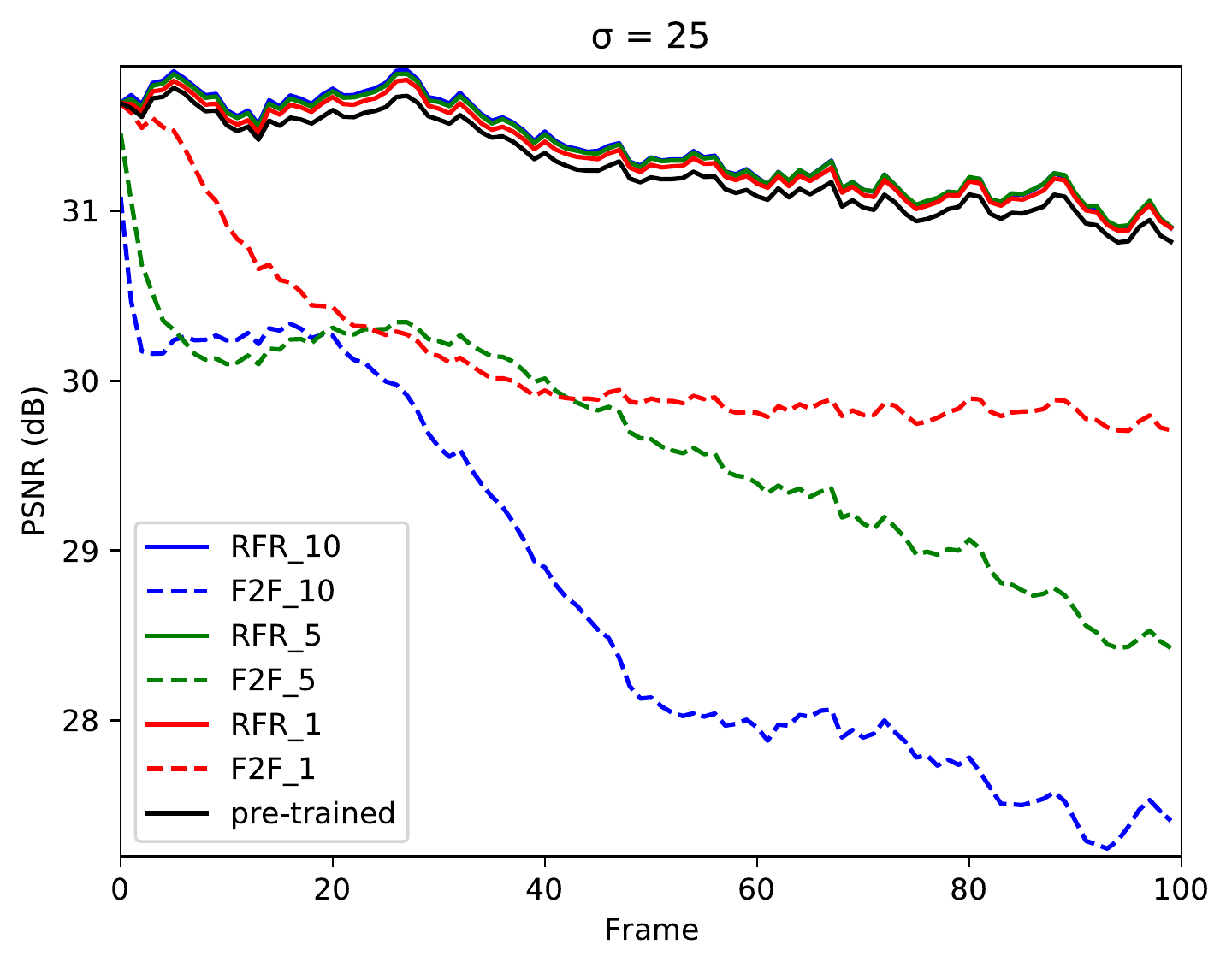}
    
    \subcaption{}
\end{minipage}
% \hspace{0.01\linewidth}
\begin{minipage}[c]{\linewidth}
\centering
    \includegraphics[width=\linewidth, height=0.5\linewidth]{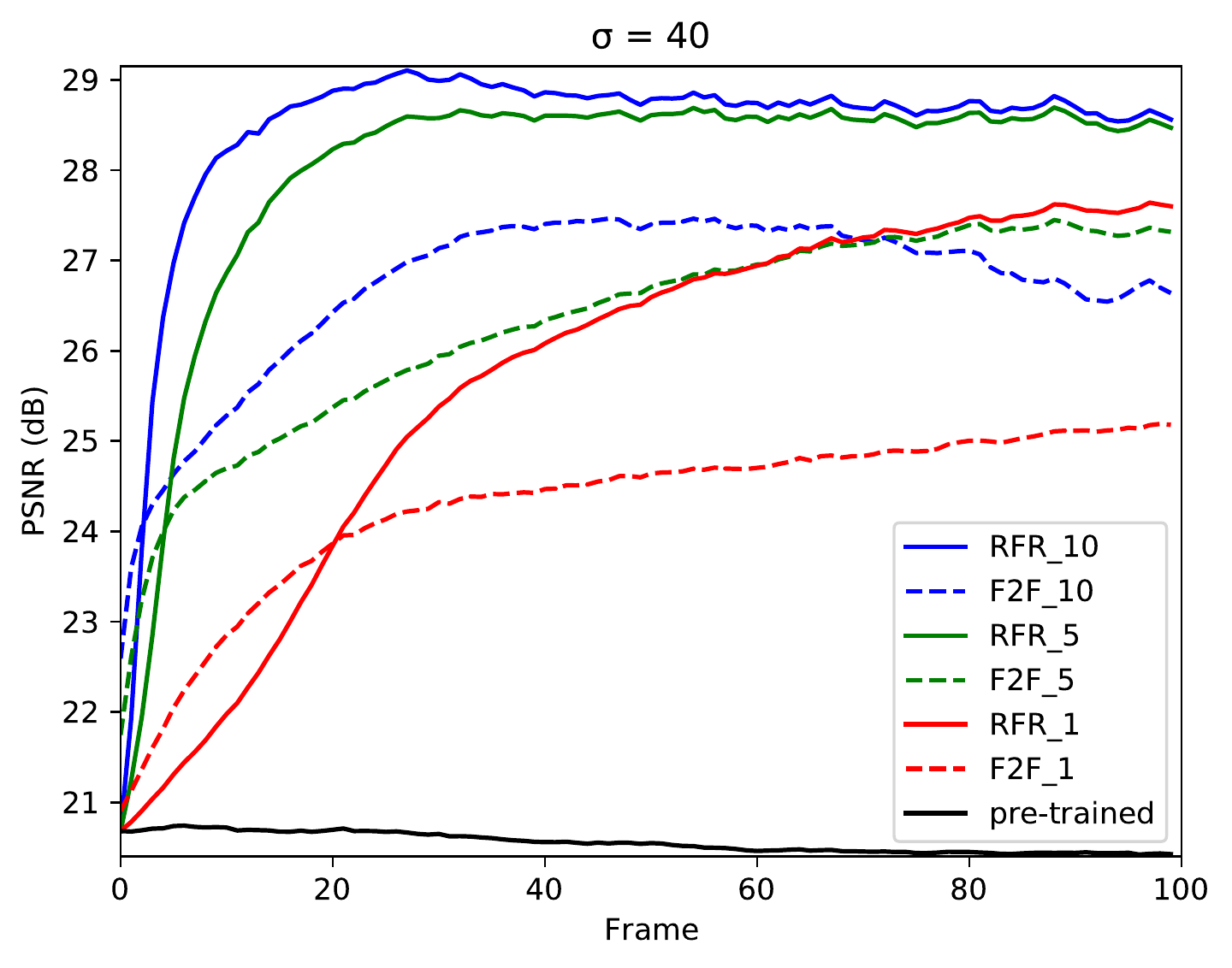}
    \subcaption{}
\end{minipage}
% \hspace{0.01\linewidth}
\begin{minipage}[c]{\linewidth}
\centering
    \includegraphics[width=\linewidth, height=0.5\linewidth]{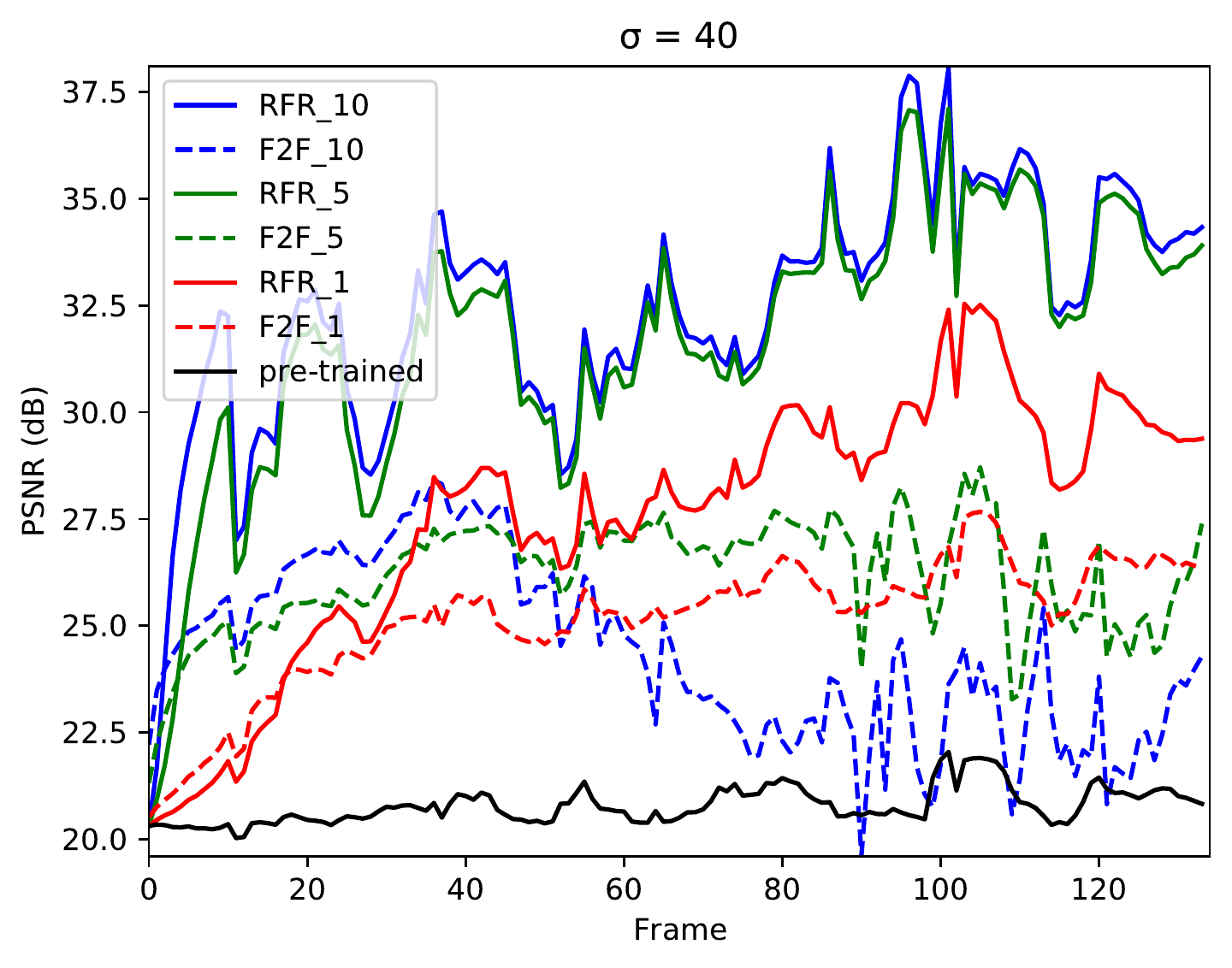}
    \subcaption{}
\end{minipage}
\caption{Comparisons of restore-from-restored (RFR) and frame-to-frame (F2F) online fine-tuning methods. The number $i$ in \qq{F2F\_$i$} or \qq{RFR\_$i$} indicates the number of updates for each frame~\cite{Ehret}. (a) Denoising results when the test input noise level is higher ($\sigma$ = 40) than that in training ($\sigma$ = 25). (b) Denoising results when noise level during the test and training stages is identical. (c) Denoising results on a video with large motions when noise level during the test and training stages is identical.}
\label{fig_comparision_F2F}
\end{figure}

We compare our online denoising method with the frame-to-frame training algorithm~\cite{Ehret}, and their official code and parameters are used for the fine-tuning.
As the baseline, a fully pre-trained DnCNN~\cite{DnCNN}, which is trained with the Gaussian random noise ($\sigma$ = 25) on the large external dataset is used, and then fine-tuned on the Derf dataset using frame-to-frame and our algorithms. For fair comparison, the same hyper-parameters (e.g., optimizer, learning rate, number of updates for each frame) are used to run the both frame-to-frame and our restoration algorithms; the comparison results are provided in~\figref{fig_comparision_F2F}.

First, we show that the frame-to-frame algorithm cannot outperform the baseline network when the noise distribution of the test video is identical to that in the large external train set (\ie, $\sigma$ = 25). By contrast, our online restoration algorithm can still elevate the denoising performance compared with the fully pre-trained baseline by adapting the network parameter to the specific input video (\figref{fig_comparision_F2F} (a)).
Next, compared with the baseline, the frame-to-frame algorithm and our approach can improve denoising quality by a large margin when the noise distribution of the input video (\ie, $\sigma$ = 40) is different from that of the training dataset (\figref{fig_comparision_F2F} (b)). 
However, when the input video includes large motion displacement, the frame-to-frame algorithm fails in estimating accurate optical flow and thus can show worse performance than the baseline; by contrast, our algorithm predicts consistently better results because ours does not rely on the optical flow (solid blue lines in~\figref{fig_comparision_F2F} (c)).

In the comparison with the frame-to-frame method, our algorithm uses additional information regarding the noise distribution of the test input but shows considerably better denoising results. By contrast, the frame-to-frame algorithm requires additional resources and longer run-time to compute optical flow among video frames.

\if
As a backbone network, DnCNN is used and A fully pre-trained parameter of DnCNN\footnote{Pytorch version. \url{https://github.com/SaoYan/DnCNN-PyTorch}}~\cite{DnCNN} used in ~\cite{Ehret}. The DnCNN is trained with Gaussian random noise ($\sigma$ = 25) on grayscale image. We convert color Derf testsets into grayscale for experiments. We set all the hyperparameters same for fair comparision (\ie, learning rate and the number of iterative update for each frame~\cite{Ehret}). The official code for online frame-to-frame training is available, so we compare it with our online restore-from-restored training. First, \figref{fig_comparision_F2F} (a) presents that both methods can effectively remove noise even if the test noise level is different from training ($\sigma$ = 40). It also shows that the fine-tuned performance with our method outperforms the fine-tuned one with frame-to-frame training. Second, we fine-tune on noise level consistent with pre-training ($\sigma$ = 25). We found that frame-to-frame training degrades the baseline performance contrary to our method when noise level is similar to pre-training so little noise exist in the denoised frame as shown in \figref{fig_comparision_F2F} (b). Lastly, we evaluate two online fine-tunig methods on a video with large movements between neighboring frames. Unlike other videos, we extract frames at every 10th time-step from the \q{\textit{factory}} video with 1339 frames in Derf datasets to make a big motion. \figref{fig_comparision_F2F} (c) shows the results on this test video. Frame-to-frame training results demonstrate that accurate optical flow and registration are difficult especially with a large motion. Our method does not have any registration step, thus, it is more computationally efficient during fine-tuning and more stable for the performance.
\fi

%-------------------------------------------------------------------------
\section{Conclusion}

In this work, we present a new training algorithm for video denoising; this algorithm is straightforward and easy to train and produces state-of-the-art denoising results. Our training approach is based on the self-supervision and thus allows the network to adapt its pre-trained parameter for the given specific input video without using ground-truth clean frames.
As we use the restored version of the input noisy frames rendered by the pre-trained denoiser as our fine-tuning target (pseudo clean images), we call the proposed algorithm \qq{restore-from-restored.}
Moreover, in contrast to conventional video restoration approaches, we restore the clean images without using accurate optical flow.
We describe how the proposed training algorithm can exploit recurring patches among input video frames and improve the denoising performance. We also demonstrate the superiority of the proposed algorithm and show considerable improvements on the various benchmark datasets.

\if
\section*{Acknowledgement}
acknowledgement

acknowledgement

acknowledgement

acknowledgement

acknowledgement
\fi

%-----------------------------------------------------------------------
% \clearpage
\FloatBarrier
{\small
\bibliographystyle{ieee_fullname}
\bibliography{egbib}
}

\end{document}